# An Overview of the Drone Open-Source Ecosystem

John Glossner[1,2], Samantha Murphy[1], and Daniel Iancu[1]

[1]Optimum Semiconductor Technologies Inc., Tarrytown, NY

[2]University of Science and Technology Beijing

**Abstract**—Unmanned aerial systems capable of beyond visual line of sight operation can be organized into a top-down hierarchy of layers including flight supervision, command and control, simulation of systems, operating systems, and physical hardware. Flight supervision includes unmanned air traffic management, flight planning, authorization, and remote identification. Command and control ensure drones can be piloted safely. Simulation of systems concerns how drones may react to different environments and how changing conditions and information provide input to a piloting system. Electronic hardware controlling drone operation is typically accessed using an operating system. Each layer in the hierarchy has an ecosystem of open-source solutions. In this brief survey we describe representative open-source examples for each level of the hierarchy.

## 1 INTRODUCTION

OPEN SOURCE software refers to software where the source code is publicly available[1]. This allows the software to be inspected, modified, and enhanced. Open-source software is generally subject to licensing terms. These can be permissive or restrictive. Permissive licenses include MIT[2] and Apache-2.0[3]. Permissive licenses generally allow for modifications in commercial applications. More restrictive licenses such as AGPL[4] require automatic patent licenses and release of any modifications making them less appropriate for certain commercial applications.

Similarly, open-source hardware is hardware whose design is made publicly available allowing anyone to study, modify, distribute, make, and sell the design or hardware based on that design[5].

Open Application Protocol Interfaces (APIs) allow proprietary software to communicate with other software without access to source code. The OpenAPI initiative within the Linux Foundation provides specifications on for defining APIs[6].

Unmanned Aerial Vehicles (UAVs) or Unmanned Aerial Systems (UASs), have a large ecosystem of open-source software. Nearly every aspect of building, flying, and navigating a drone can be implemented using open-source technologies.

UASs can be organized into a top-down hierarchy of layers that include 1) Flight supervision, 2) UAV command and control, 3) Simulation, and 4) Operating Systems, and 5) Hardware.

Flight management is concerned with traffic management, flight authorization, proof of safety, and remote identification. With an anticipated 3-4 million Small UAVs (sUAV), colloquially referred to as drones, will need to have automated processes for safe flight [1].

UAV command and control concerns the hardware and software used to fly and control UAVs. This includes messages sent from a controller to the UAV either in the form of manual (joystick) inputs and autopilots. Open API's such as MAVLINK have been developed to allow UAVs operating with proprietary software to be controlled by other proprietary or open-source software.

UAV simulation concerns all aspects of UAV flight and may include aspects of path management, vehicle models, physics models, weather models, and visualization. With respect to UAV wireless networking, simulation may include communications systems such as LTE, antenna patterns, channel models, and path loss.

Hardware and operating systems interface UAV software with the physical UAV. Specialized open-source operating systems have been developed to control UAVs running on open hardware.

In this survey we look at a sampling of open-source projects used in UAV applications. Section 2 addresses open-source systems for flight supervision including Unmanned Traffic Management (UTM), UAS flight management, flight safety and authorization, and remote identification. Section 3 highlights open API's and open-source autopilots for UAS command and control. Section 4 considers open-source simulators capable of flight simulation, network traffic modeling, and communications system modeling including interference. Section 5 discusses open-source operating systems. Section 6 describes open-source hardware. Finally, in Section 7, we present concluding remarks.

## 2 FLIGHT SUPERVISION

In this section we consider open-source systems used for flight supervision including Unmanned Traffic Management (UTM), UAS flight management, flight safety and authorization, and remote identification.

---

[1] https://opensource.org/docs/osd
[2] https://opensource.org/licenses/MIT
[3] https://opensource.org/licenses/Apache-2.0
[4] https://www.gnu.org/licenses/agpl-3.0.en.html
[5] https://www.oshwa.org/definition/
[6] https://spec.openapis.org/oas/latest.html

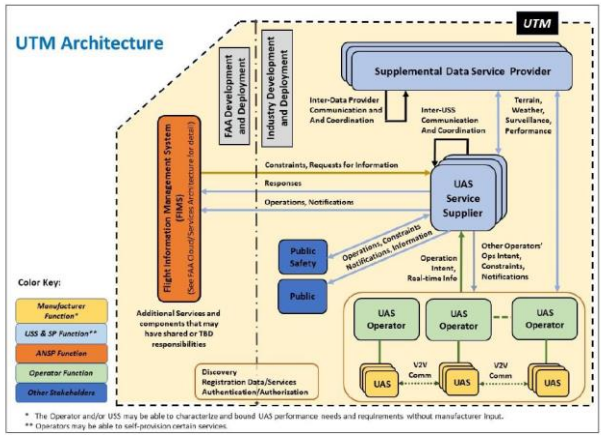

Fig. 1. UTM Architecture

## 2.1 UTM

In March of 2020 the Federal Aviation Administration (FAA) in collaboration with the National Aeronautics and Space Administration (NASA) and other federal agencies published revision 2.0 of the Unmanned Aircraft Systems (UAS) Traffic Management (UTM) Concepts of Operations.[7] This collaborative effort describes more detailed UTM protocols aimed at enabling multiple, beyond visual line-of-sight (BVLOS) drone operations within the same airspace.

Fig. 1. shows and overview of a notional UTM architecture. Operators of manned aircraft are not required to participate in UTM but can voluntarily participate. UAS visual-line-of-site operators are not required to use data exchanges but may use UTM services to meet regulatory and policy requirements. UAS BVLOS aircraft must use UTM services and possibly Detect and Avoid (DAA) equipment.

UAS Service Supplier (USS) examples include operations planning, flight intent sharing, conflict identification, conformance monitoring, remote identification (RID), airspace authorization, airspace management, and discovery of other USS services. UAS Service Suppliers may use data from Public and Public Safety sources when supplying UAS operators with data. UAS Supplemental Data Service Providers (SDSP) may include terrain/obstacle data, weather, and surveillance.

All USS services are anticipated to be from industry development and deployment. As part of the UTM architecture, the FAA develops and deploys a Flight Information and Management System (FIMS). The FIMS makes real-time airspace constraints available of UAS operators through USS services.

### InterUSS Platform

The InterUSS Project is hosted by the Linux Foundation and developed by federal and industry participants including Wing, Uber, and Airmap. It is Apache 2.0 licensed with code available on github[8]. It enables trusted, secure, and scalable interoperability between USSs furthering safe, equitable and efficient drone operations without requiring personally identifiable information[9].

InterUSS performs two primary functions: 1) Discovery of other USSs that it needs to obtain data from, and 2) Synchronization Validation supporting deconfliction and ensuring drone operations have complete and up-to-date information regarding constraints and other operations in the area.

The Discovery and Synchronization Service (DSS) is a federated/distributed concept facilitating the discovery of relevant airspace data and synchronization between multiple participants when updating airspace data. All discoverable data is associated with an "Entity". It is tracked and communicated by the DSS which is characterized by a 4D volume (spatial and time bounds). Entities can be created in the DSS using Entity-type-specific APIs[10].

After an Entity is created in the DSS, it will be discovered by any other USS inquiring about any other Entity that intersects the new Entity's 4D (space-time) volume. Existing USSs with previously declared interests in any airspace-time will receive notification of an intersecting Entity's 4D volume and its details via the Subscription notifications mechanism.

### Global UTM Association

The Global UTM Association (GUTMA)[11] is a non-profit consortium of worldwide Unmanned Aircraft Systems Traffic Management (UTM) stakeholders. Its purpose is to foster the safe, secure, and efficient integration of drones in national airspace systems. Its mission is to support and accelerate the transparent implementation of globally interoperable UTM systems. GUTMA members collaborate remotely. GUTMA has released several Apache 2.0 repositories on github last updated in 2018 and 2019.

GUTMA's Flight Declaration Protocol[12] targets drone operators and allows those operators to digitally share interoperable flight and mission plans. The Flight Logging Protocol[13] targets drone manufacturers and USSs. It offers an interoperable interface to access post-flight data. In the future, it aims at enabling access to inflight telemetry data. The Air Traffic Data Protocol[14] aims to standardize the transmission of sensor data to apps and services. The Drone Registry Database Schema[15] and Brokerage API[16] provides an interoperable drone registry.

## 2.2 Flight Management

A Ground Control Station (GCS) is a software application running on a ground-based computer that communicates with a UAV. Flight management controls are sent to a UAV from the GCS. GCSs typically provide for path planning, mapping, and real-time flight statistics superimposed on a map.

---

[7] https://www.faa.gov/uas/research_development/traffic_management/
[8] https://github.com/interuss/dss
[9] https://interussplatform.org/
[10] https://github.com/interuss/dss/blob/master/concepts.md
[11] https://gutma.org/
[12] https://github.com/gutma-org/flight-declaration-protocol-development
[13] https://github.com/gutma-org/flight-logging-protocol-development
[14] https://github.com/gutma-org/airtraffic-data-protocol-development
[15] https://github.com/gutma-org/droneregistry
[16] https://github.com/gutma-org/droneregistry-broker

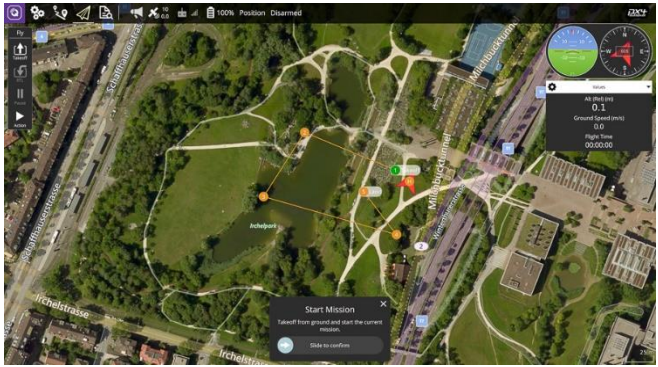

Fig. 2. QGroundControl Visualization Flight Planning Software

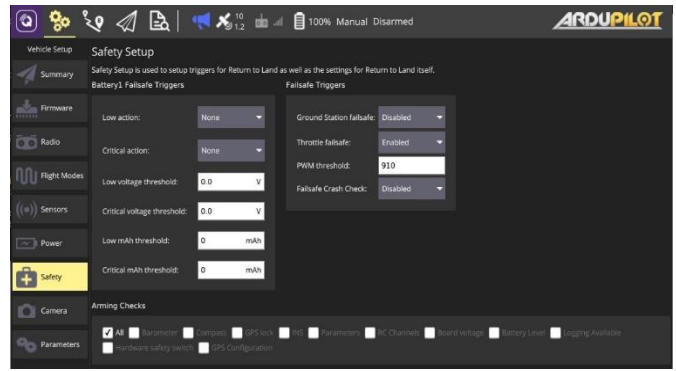

Fig. 3. ArduPilot safety setup triggers for Return to Launch

*QGroundControl*

Fig. 2. Shows a QGroundControl (QGC) screenshot. QGC is a software ground control station providing full flight control and vehicle setup for PX4 or ArduPilot vehicles. The Dronecode Foundation, a vendor-neutral sub-group of the Linux Foundation, manages and develops the dual-licensed GPLv3 and Apache 2.0 code[17].

The software is written using Qt QML[18] providing for both JavaScript and C++ development. It is designed to provide a cross-platform single codebase with a tablet style user interface.

Using the MAVLink protocol, QGC assists operators by supporting mission planning activities during autonomous flights. The flight map display shows position, flight track, waypoints, and drone instruments. Video can be overlayed on the flight map.

Flight planning is provided with geofence, rally points, and complex patterns. Geofencing allows the creation of virtual regions by establishing geographical barriers. Within those barriers, the operator either allows or disallows flight. If the UAV violates the established barriers, the operator is provided the opportunity to take action..

Rally points are an alternative landing or loiter location. In the event Return to Home or Return to Launch mode proves unsafe, difficult, or impossible, operators can utilize Rally Points.

A pattern tool is provided for complex flight patterns. It supports: 1) surveys as grid flight plans over a polygonal area, 2) structural scans as a grid flight pattern over circular or polygonal vertical surfaces, 3) corridor scans that follow a polyline which can be used to survey roads, and 4) landing patterns for fixed wing vehicles.

*MAVProxy*

ArduPilot maintains and develops a command line ground stations software package called MAVProxy[19]. It is licensed under GPLv3.

MavProxy supports the MAVLink protocol, and includes a convenient syntax for generating MAVLink commands, thereby acting as a fully functional GCS for a UAV during its autonomous flight. Its key advantage is its usefulness in companion computing which is achieved by providing the capability of forwarding messages via UDP from a UAV to multiple GCS on other devices.

While primarily a command-line, console-based application, plugins are available providing basic GUI. It runs on any POSIX OS with python (Linux, OS X, Windows, etc.).

*Others*

APM Planner 2.0[20] is an open-source ground station application for MAVlink based autopilots including ArduPilot Mega[21] (APM) and PX4/Pixhawk that can be run on Windows, Mac OSX, and Linux.

Mission Planner is a Windows-only ground station application for the ArduPilot open-source autopilot project[22].

Tower is a GPLv3 licensed Ground Control Station (GCS) Android app built atop DroneKit-Android, for UAVs running Ardupilot software[23].

### 2.3 Flight Safety / Authorization

The Pilot-in-Control is responsible for ensuring the safe flight of any UAS. Autopilots, such as the software from QGroundControl, provide several, and often configurable safety features including failsafe actions for low battery, loss of remote-control signal, data link failures, and geofence violations. Fig 3. shows the ArduPilot failsafe settings that trigger a return to launch.

*LAANC API's*

Drone pilots planning to fly under 400 feet in controlled airspace around airports must receive FAA airspace authorization before they fly. The UAS Low Altitude Authorization and Notification Capability (LAANC) is available to pilots operating under the Small UAS Rule Part 107, or under the exception for Recreational Flyers[24]. As of this writing more than 700 airports support LAANC[25] and seven LAANC UAS Service Suppliers[26].

Access LAANC is provided through one of the FAA approved USS LAANC provides: 1) drone pilots access to controlled airspace at or below 400 feet, 2) awareness of where pilots can and cannot fly, and 3) Air Traffic Professionals with visibility into where and when drones are operating.

---

[17] https://dev.qgroundcontrol.com/master/en/index.html
[18] https://doc.qt.io/qt-5/qtqml-index.html
[19] https://ardupilot.org/mavproxy/
[20] https://ardupilot.org/planner2/
[21] https://www.ardupilot.co.uk/
[22] https://ardupilot.org/planner/docs/mission-planner-overview.html
[23] https://github.com/DroidPlanner/Tower
[24] https://www.faa.gov/uas/programs_partnerships/data_exchange/
[25] https://www.faa.gov/uas/programs_partnerships/data_exchange/laanc_facilities/
[26] https://www.faa.gov/uas/programs_partnerships/data_exchange/

The FAA published a concept of operations[27], USS performance rules[28], and the API both as a document and on their github site[29].

*Google Wing OpenSky*

While not open-source software, OpenSky[30] is a free mobile and web application providing certain airspace information, i.e., when and where it is safe to fly, to drone operators for safe autonomous navigation of the sky. OpenSky supports all types of drone users including recreational or commercial. OpenSky is currently available in the United States and Australia.

OpenSky is approved by the FAA for Part 107 and 44809 UAS Operations in the United States[31]. LAANC submissions can be performed in-app.

### 2.4 Remote Identification (RID)

Remote Identification (RID) is a unique electronic identifier for UAVs whose purpose is comparable to a license plate for a ground-vehicle. Direct broadcast of the identifier limits reception to anyone within range of the signal. Network publishing of RID by a USS is also possible. Direct broadcasting of RID will be required on all UASs beginning in 2023[32]. Recreational pilots with UASs under 0.55 pounds are exempt from RID.

*Altitude Angel*

Altitude Angel's Scout[33] is an Apache 2.0 licensed open-source hardware and firmware using cellular communications to send position reports to and receive navigation assistance from a UTM provider.

Primarily intended for use in commercial and industrial drone applications, Scout provides the capability to securely obtain and broadcast a form of network remote ID, widely seen as a necessary step for enabling routine drone use and flights beyond visual line of sight[34].

Altitude Angel also has made available a surveillance API[35] allowing integrators to both share and receive, in near real time, flight data from a variety of sensors and devices, providing a comprehensive real-time picture of the airspace.

*Open Drone ID*

Open Drone ID[36] is a project providing a low cost and reliable "beacon" capability for drone identification when the drone is within the range of a receiver. The current specification is based on the American Society for Testing Materials (ASTM) Remote ID standard[37]. It includes Bluetooth\*

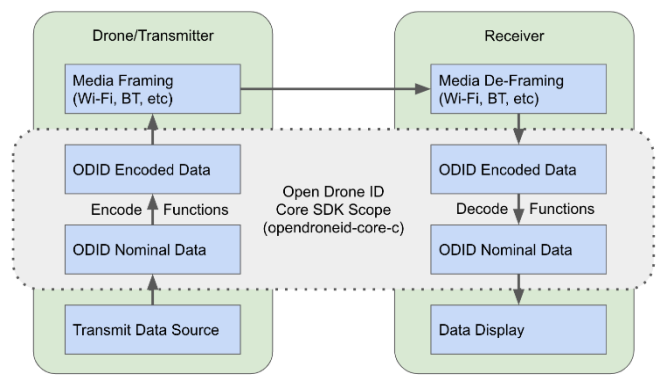

Fig. 4. Open Drone ID Architecture

Legacy Broadcasts Packets (Advertisements), the new Bluetooth 5 (long range) Advertising Extensions, and WiFi implementation advertisements based on Neighbor Awareness Network protocol. Network access APIs are under development.

Fig. 4 shows the Open Drone ID Architecture and the focus of the core Software Developers Kit (SDK). The data transmission by the drone is encoded and sent over a wireless link to a receiver listening on that link. Once the transmission is received, the data is decoded for display.

Open Drone ID provides an Apache 2.0 licensed function library for encoding and decoding (packing/unpacking) ASTM Remote ID messages[38]. The code is also compatible with the upcoming European ASD-STAN Direct Remote ID standard. MAVLink messages for drone ID are also available[39].

## 3 COMMAND AND CONTROL

Command and control are the processes of controlling UAV flight by sending signals (commands) to the UAV. The origin of the signal may be manual via a human operator or computer-generated via an auto-pilot.

The control signal is sent to the UAV through either proprietary protocols or open APIs. In this section we review common open APIs and autopilots.

### 3.1 API's

APIs provide a mechanism for software software-to-software communications without the need to use source code.

*MAVLINK*

The Micro Air Vehicle Communication Protocol (MAVLink)[40] is a lightweight, header-only message marshaling library for micro air vehicles. It is widely used for communicating commands and telemetry between ground stations and autopilots.

---

The Dronecode Foundation provides a Software Development Kit (SDK) primarily written in C++ for MAVLINK along with bindings for common languages such as Python, Java, Rust, etc.

MAVLINK defines commands that can be sent to any system supporting the MAVLINK API including UAVs. Typical commands and information sharing include: 1) Mode: Preflight, Manual, Guided (waypoints), Autonomous, 2) Sensor Status: gyro, accelerometer, computer vision, 3) Mission Type: geofence, rally, 4) UTM Flight State: ground, airborne, emergency, and 5) Navigation: takeoff, return to launch, waypoints, follow, land.

In addition to typical commands there is extensive error reporting and support for specific sensors such as cameras, gimbals, etc. Koubaa published a recent survey of MAVLinK capability [2].

### PARROT

Parrot, a French-based company, makes ANAFI UAVs utilized by emergency response applications and several commercial applications, some of which include thermal imaging, inspection, mapping, etc.

They provide an SDK[41] under the very permissive BSD-3 license[42]. No registration is required to download the SDK. Whenever possible, Parrot uses standard protocols. An example of this is direct support for MavLink in their ground API.

The ground API is a ground control station for mobile devices. All features of the aircraft, including instrument telemetry, information, peripheral status (camera, gamepad, Wi-Fi), and piloting information, i.e., tracking, are made accessible through the API.

Olympe, a Python controller programming interface is also provided by Parrot. It is part of the Ground API supporting both simulated and physical drone flight.

Parrot uses its ARSDK XML message passing interface for transferring information in both the Ground API and the Olympe API. Every aspect of interfacing their APIs is available in ARSDK all the way down to the state of any LEDs on the drone.

Parrot, as part of their SDK, provides a Simulation Description Format (SDF) originally designed for Gazebo. SDF supports OpenStreetMaps, provides a structured format for retrieving drone information such as air pressure, atmospheric models, inertial measurement units, lidar information, axis of rotation, trajectory, waypoints, and is licensed under the permissive Apache 2.0 license.

## 3.2 Autopilots

Autopilots are flight control software for drones and other unmanned vehicles. They are the "brain" of the UAV, consisting of a flight software stack running on flight controller hardware.

### PX4

PX4 is an autopilot flight stack for drones. The Dronecode Foundation maintains and develops the permissibly licensed BSD-2 autopilot. The project was founded in 2012 with C/C++ code available on github[43].

PX4 supports different types of drone frames including multicopters (multiple propellers), fixed wing, and Vertical Takeoff and Landing (VTOL) vehicles.

PX4 also supports Unmanned Vehicles (UV) beyond aerial systems including Unmanned Ground Vehicles (UGV), Unmanned Surface Vehicles (USV) (e.g., boats) and Unmanned Under Water Vehicles (UUV).

A wide range of commercially available drones directly support PX4[44]. Of note, PX4 runs on most of the Pixhawk open-source hardware platforms[45]. Pixhawk-based and other flight controllers such as those based on Raspberry Pi are also available[46].

PX4 is configured using Dronecode's QGroundControl ground control station. QGC is used to load (flash) PX4 onto vehicle control hardware.

A Radio Control (RC) unit manually controls the drone. A guide to selecting an RC system is provided[47]. An RC unit is not required for autonomous flights.

PX4 uses MAVLink to communicate with the flight controller. PX4 also supports Real Time Publish Subscribe protocol, a Data Distribution Service (RTPS/DDS). It is not intended to replace MAVLink but can be used to reliably share time-critical/real-time information between the flight controller and offboard components.

Prior to flight, PX4 provides vehicle status notifications (e.g., LED, Buzzer) for flight readiness. Preflight readiness checks include proper calibration of sensors, has an SD card, has position lock, and is armed.

During flight PX4 reports aircraft state and sensor data. Several different modes are provided depending on vehicle type. For example, multicopters can be placed in Manual-Position, Manual-Altitude, Manual-Stabilized, Manual-Acrobatic, or a number of Autonomous modes including Mission, Takeoff, Land, Return, and Follow me. An Offboard autonomous mode allows a remote computer to send position, velocity, or altitude setpoints using MAVLink.

During flight PX4 uses sensors to determine vehicle state needed to maintain stabilization or enable autonomous operation. Minimum requirements include gyroscope, accelerometer, magnetometer, and barometer. A GPS is required for automatic modes.

PX4 produces logs during flights. Flight logs can be analyzed using flight log analysis tools such as Flight Review, pyulog, or PX4Tools. An extensive list of log analysis tools is maintained by PX4[48].

### ArduPilot

The ArduPilot Project[49] provides an advanced, full-featured and reliable open-source autopilot software system.

---

[41] https://github.com/Parrot-Developers
[42] https://opensource.org/licenses/BSD-3-Clause
[43] https://github.com/PX4
[44] https://px4.io/ecosystem/commercial-systems/
[45] https://px4.io/autopilots/

[46] https://docs.px4.io/master/en/getting_started/flight_controller_selection.html
[47] https://docs.px4.io/master/en/getting_started/rc_transmitter_receiver.html
[48] https://docs.px4.io/master/en/log/flight_log_analysis.html
[49] https://ardupilot.org/index.php/about

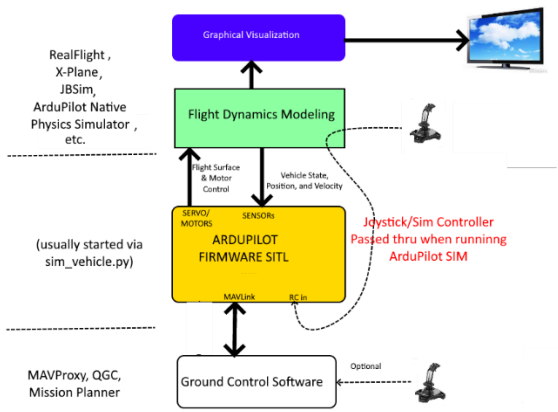

Fig. 5. System Simulation using ArduPilot

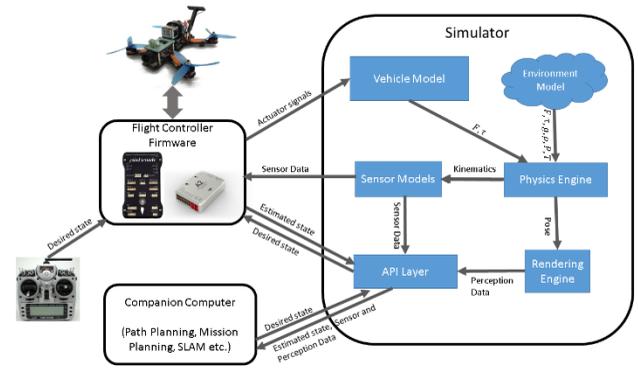

Fig. 6. AirSim System Simulation

The first ArduPilot open code repository was created in 2009. It can control many vehicle systems including conventional and VTOL airplanes, gliders, multirotors, helicopters, sailboats, powered boats, submarines, ground vehicles and even Balance-Bots. It is installed on more than a million vehicles and runs on a large number of open hardware flight controllers including Pixhawk[50].

ArduPilot's C/C++ code is distributed under the GPLv3 open-source license. A rationale for this license is described on their website[51].

ArduPilot features include: 1) fully autonomous, semi-autonomous, and fully manual flight modes with multiple stabilization options, 2) programmable missions with 3D waypoints and optional geofencing, 3) many supported sensors with communication over many different buses, 4) Fail safes and support for navigation in GPS denied environments.

ArduPilot supports the MAVLink protocol for communication with Ground Stations and Companion Computers[52]. Mission commands are stored on the flight controller in eeprom and executed one-by-one when the vehicle is switched into Auto mode.

ArduPilot has hundreds of parameters allowing the operator to configure many aspects of how the vehicle flies/drives. These can be retrieved by a ground station or companion computer.

### Paparazzi

Paparazzi UAV[53], is a GPLv2 licensed open-source drone hardware and software project encompassing autopilot systems and ground station software for multicopters/multirotors, fixed-wing, helicopters, and hybrid aircraft. It was founded in 2003 and designed with autonomous flight as the primary focus and manual flying as the secondary. From the beginning it was designed with portability and the ability to control multiple aircraft within the same system. Code is available on github[54].

## 4 SIMULATION

Many simulators have been developed that are capable of simulating aerial flights, modeling network traffic, and modeling communications systems. In this section we look at some commonly used open-source projects.

### 4.1 Flight Modeling

Flight modeling allows a UAS to fly without hardware. All aspects of the system are typically supported. Often graphical user interfaces provide detailed modeling of all aspects of the flight including sky and scenery.

### ArduPilot SITL

Fig 5. shows how ArduPilot, as Software-In-The-Loop (SITL), integrates with a simulation system. The Flight Dynamics Modeling (green box) is the simulator receiving commands from the ArduPilot Firmware SITL. This allows the vehicle to be run without any hardware.

Sensor data is sourced from a flight dynamics model in a flight simulator. ArduPilot has a number of vehicle simulators built-in including multirotor aircraft, fixed wing aircraft, ground, and underwater vehicles[55].

External simulators, particularly Gazebo[56] and AirSim[57] (described below) can interface with ArduPilot's SITL. A Gazebo plugin for ArduPilot is available as is a Gazebo with Robot Operating System (ROS) support.

### AIRSIM

AirSim is a Microsoft Research developed simulator for drones and ground vehicles [3]. It is used by NASA and several research universities[58]. The code is released under the permissive MIT License. It is built on Unreal Engine with an Unreal plugin available[59] [4]. Unreal Engine is not open source. It is proprietary software with source code access under licensing restrictions[60].

The goal is to develop a platform for AI research. Therefore, AirSim exposes APIs to retrieve data and control vehicles in a platform independent way.

Fig. 6 shows an AirSim simulation system. Manual commands may be sent to Flight Controller Firmware though the use of a Radio Controller (RC). AirSim also allows programmatic control by exposing APIs through Remote Procedure Calls (RPC) assessable to a number of languages including C++, Python, C#, and Java. These APIs are part of a separate independent cross-platform enabling deployment on a companion computer.

---

[50] https://ardupilot.org/copter/docs/common-autopilots.html
[51] https://ardupilot.org/dev/docs/license-gplv3.html
[52] https://ardupilot.org/dev/docs/companion-computers.html
[53] https://wiki.paparazziuav.org/wiki/Main_Page
[54] https://github.com/paparazzi/paparazzi
[55] https://ardupilot.org/dev/docs/sitl-simulator-software-in-the-loop.html
[56] https://ardupilot.org/dev/docs/using-gazebo-simulator-with-sitl.html
[57] https://ardupilot.org/dev/docs/sitl-with-airsim.html
[58] https://microsoft.github.io/AirSim/who_is_using/
[59] https://microsoft.github.io/AirSim/
[60] https://www.unrealengine.com/en-US/eula/publishing

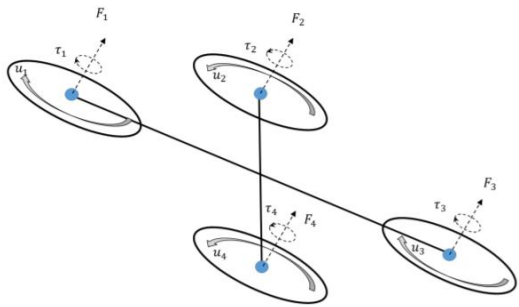

Fig. 7. AirSim Drone Vehicle Model

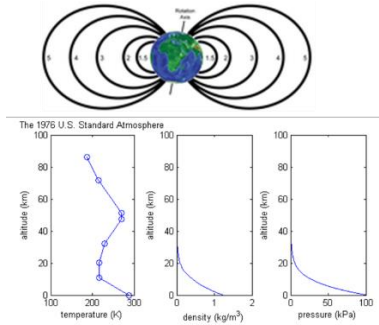

Fig. 8. AirSim Magnetic and Atmospheric Environmental Models

A special computer vision mode is provided that inhibits the physics engine but allows the use of the keyboard to move around the scene collecting camera data including pose, depth, disparity, surface normal, and object segmentation.

Commands, whether manually or programmatically communicated, use Microsoft's MIT Licensed MavLinkCom library. It is a cross-platform C++ library for any MAVLink vehicle but specifically works well with PX4 based drones[61].

The simulator is modular and supports 1) vehicle models, 2) environment models, 3) physics engines, 4) sensor models, 5) rendering engine, and 6) APIs to communicate with the simulator.

Fig. 7 shows AirSim's drone vehicle model. Multiple vehicle models are simultaneously supported allowing for example drone swarms to be simulated[62]. Vehicles are treated as rigid bodies with an arbitrary number of actuators generating force and torque. Parameters for the vehicle model include mass, inertia, linear/angular drive, friction, and restitution[63]. Forces are assumed to be generated in the normal direction. Positions and normals may change during simulation to allow for VTOL simulation.

Environment modeling includes support for gravity using a binomial first order expansion of Newton's Law of Universal Gravitation. Fig. 8 shows AirSim's magnetic and atmospheric models. Magnetic fields are modeled using the NOAA world magnetic model. Atmospheric models of temperature, density, and pressure are based on distance. For heights below 51km (troposphere/stratosphere) the 1976 U.S. standard atmospheric model is used (constant lookup tables). For heights from 51km to 86km (mesosphere) an equation with the only parameter being height is used. For heights greater than 86km (thermosphere) air density is computed based on pressure and temperature[64].

AirSim uses Unreal Engine's physics engine, Physx, for ground vehicle simulation but uses a custom kinematic (rigid body) physics engine called FastPhysics for multirotor drone modeling[65]. FastPhysics uses six parameters: position, quaternion-based orientation, linear velocity, linear acceleration, angular velocity, and angular acceleration. Collisions are supported using Unreal Engine to provide impact position, impact normal, and penetration depth.

AirSim provides a weather API. For ground vehicles Unreal Engine supports weather modeling if materials are applied to the scene. FastPhysics does not appear to support weather modeling.

AirSim supports a variety of sensors including gyroscopes, accelerometers, barometers, magnetometers, GPS including latency simulation, and Lidar.

AirSim uses Unreal Engines for rendering. Visual near photo-realistic rendering is accomplished using physically based materials, photometric lights, planar reflections, ray-traced distance field shadows, and lit translucency.

*FlightGear*

FlightGear is a GPLv3 licensed open-source flight simulator[66] running on multiple platforms including Windows, Linux, and Mac. The goal of the FlightGear project is to create a sophisticated and open flight simulator framework for use in research or academic environments, pilot training, as an industry engineering tool, for DIY-ers to pursue their favorite interesting flight simulation ideas.

It supports fully configurable aerodynamics and a propulsion system capable of modeling the complex flight dynamics of an aircraft.

FlightGear provides choices between three primary Flight Dynamics Models[67]: 1) JSBSim[68], a generic, 6 degrees of freedom, flight dynamics model for simulating motion of flight vehicles. Rotational earth effects are also modeled[69], 2) YASim[70], an integrated part of FlightGear, simulates the effects of the airflow on the different parts of an aircraft, and 3) UIUC[71], based on LaRCsim originally written by the NASA, it extends the code by allowing aircraft configuration files and by adding code, allowing simulation of aircraft under icing conditions.

---

Key features of FlightGear include 1) extensive and accurate world scenery database with over 20,000 real world airports, terrain at 3 arc second resolution, and day and night lighting, 2) detailed sky model with accurate time of day and correct real-time placement of the sun, moon, stars, 3) flexible and open aircraft modeling system with realistic instrument behaviors, and 4) networking options to communicate with GPS receivers, external autopilots, and other software.

### GAZEBO / Ignition

Gazebo, a dynamic 3D multi-robot simulator, is able to simulate populations of robots accurately and efficiently in complex indoor and outdoor environments [5][6]. Gazebo is particularly useful for defining indoor flights or swarms. Gazebo is license under the permissive Apache 2.0 open-source license[72]. It is maintained by the Open Robotics Foundation[73] and has been used in several technology challenges sponsored by DARPA, NIST, NASA, and Toyota[74].

Similar to AirSim, Gazebo's simulation architecture supports vehicle models, environment models, physics engines, sensor models, and APIs to communicate with the simulator.

Robot models provided[75] include TurtleBot, and iRobot Create. Parrot drones can use Gazebo with the Parrot Sphinx simulation tool[76]. The Parrot Bebop2 model is available directly from Gazebo[77]. Specific quadcopter models are also available for use with the PX4 autopilot[78]. Gazebo provides a Simulation Description Format (SDFormat) to design your own robot model[79].

Gazebo supports multiple physics engines including the Open Dynamics Engine (ODE)[80], Bullet[81], Simbody[82], and DART[83]. These physics engines have permissive licenses suitable for commercial uses.

The ODE physics engine is often used. It is an industrial quality library for simulating rigid body dynamics. It includes built-in collision detection. It is dual-licensed under LGPL and BSD licenses.

Gazebo supports many sensors including altimeters, IMUs, GPS, magnetometers, lidars, cameras, depth cameras, multi-cameras, and wide-angle cameras. Even RFID tags and wireless transceivers are supported[84]. Noise models can optionally be applied to the sensors.

Rendering is accomplished using the MIT Licensed OGRE[85]. It supports all major platforms including Windows, Linux, Max OSX, Android, and iOS. Supported rendering APIs include Direct3D, Metal, OpenGL, and WebGL. A small sampling of supported features include:[86] 1) materials with level of detail (LOD), 2) shaders using HLSL, GLSL, and SPIRV, 3) progressive meshes with LOD, 4) skeletal animation with multiple bone weight skinning and manual bone control, 5) flexible scene management using plugins, 6) multiple shadow rendering techniques, and 7) special effects including particle systems, skyboxes, and billboarding.

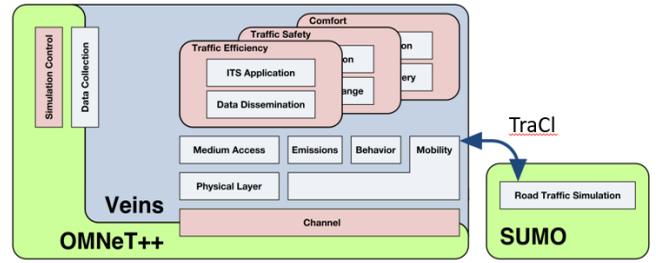

Fig. 9. Veins Architecture Overview

Gazebo's API[87] allows for communicating with the simulator and facilitates easy development of plugins. Simulations can run locally on remote servers using socket-based message passing, or in the cloud using CloudSim[88].

The latest release of Gazebo (v11), released in May 2019, added physically based rendering (PBR) materials, air pressure and RGBD sensors, a global wind model, direct support for UAVs, and battery models[89].

Open Robotics is now focusing on Gazebo's successor multi-robot simulator called Ignition. Both use SDFormat as vehicle inputs. Currently, Ignition is well behind Gazebo 11's feature support[90].

### 4.2 Traffic Modeling

Traffic modeling is concerned with simulating real-world traffic interactions. Typically, this means terrestrial traffic, but simulation of aerial traffic is also possible.

### VEINS / Sumo / OMNet++

Fig. 9 shows The Vehicle Network Simulator (Veins) architecture. It is a GPL v2 or later open-source framework for running vehicular network simulations. Source code is available on github[91]. It is based on two well-established simulators: OMNeT++ and SUMO.

Network simulation in Veins is performed by the Objective Modular Network Testbed in C++ (OMNeT++). It is an extensible, modular, component-based C++ simulation library and framework, primarily for building network simulators [10]. OMNeT++ source code is available on github[92] under an academic license[93] but requires a commercial use license for non-academic use.

Network simulation is meant in a broader sense to include wired and wireless communication networks, on-chip networks, queueing networks, etc. Domain-specific functionality such as support for sensor networks, wireless ad-hoc networks, Internet protocols, performance modeling, photonic networks, etc., is provided by model

---

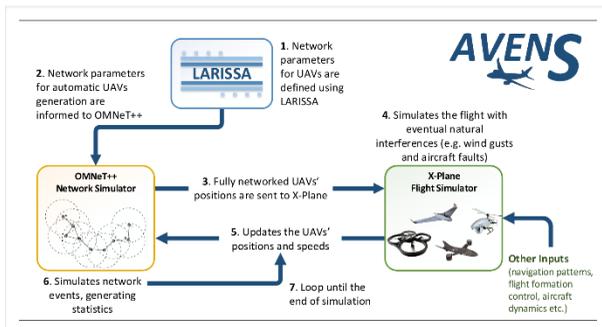

Fig. 10. AVENS Architecture Overview

frameworks, developed as independent projects[94].

Road traffic simulation in Veins is performed by Simulation of Urban Mobility (SUMO)[95]. SUMO is an Eclipse Public License Version 2[96] open source, highly portable, microscopic traffic simulation package designed to handle large road networks and different modes of transportation. It is developed by employees of the Institute of Transportation Systems at the German Aerospace Center.

Veins extends OMNeT++ and SUMO offering a comprehensive suite of models for Inter-Vehicular Communication (IVC) simulation [8]. Veins features include 1) online re-configuration and re-routing of vehicles in reaction to network packets, 2) trusted vehicular mobility model by the Transportation and Traffic Science community, 3) fully-detailed models of IEEE 802.11p and IEEE 1609.4 DSRC/WAVE network layers, 4) models for ETSI ITS-G5, 4G/5G cellular networking like 3gpp LTE and C-V2X as part of the a the INET subproject Framework, 5) import whole scenarios from OpenStreetMap, including buildings, speed limits, lane counts, traffic lights, access and turn restrictions, and 6) models of shadowing effects caused by buildings as well as by vehicles. A full list of features is available at the Veins website[97].

### AVENS

Veins is primarily intended for ground vehicles. The Ariel Vehicle Network Simulator (AVENS) has extended Veins to drones[9]. Fig. 10 shows the AVENS architecture. Network parameters for UAVs are defined using their Layered Architecture Model for Interconnection of Systems in UAS (LARISSA). Network parameters for OMNeT++ are then automatically generated. OMNeT++ sends multiple drone positions to the closed source X-Plane flight simulator[98]. X-Plane simulates environmental conditions (wind gusts, etc.) sends updated positions and speed to OMNeT++. OMNeT++ simulates network events and generates statistics. Source code is available on github[99] licensed under GPLv3. There is a video overview on their website[100].

## 4.3 Veins Wireless Communications Modeling

Veins' modules provide access to OMNeT++ wired and wireless networking modules. In this section we describe some relevant modules. Other module descriptions such as IEEE 802.11p[101], obstacle shadowing[102], antenna pattern modeling[103], and even machine learning applications to vehicular networking[104] are available at the Veins website[105].

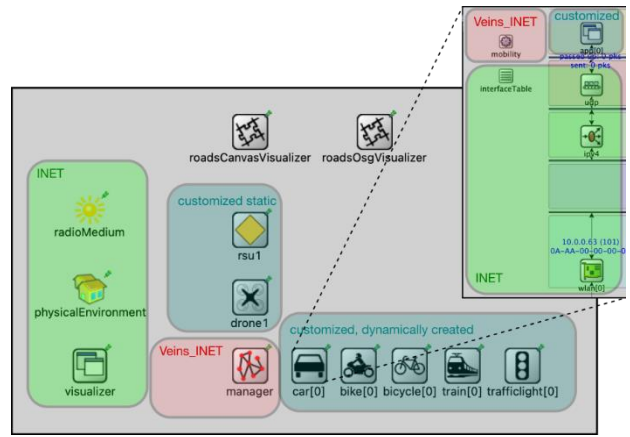

Fig. 11. Veins INET Simulation Modules

### Veins Physical Layer Modeling

The physical layer modeling toolkit[106] MiXiM incorporated into Veins makes it possible to employ accurate models for radio interference, as well as shadowing by static and moving obstacles [11]. It can be used to evaluate the effect of different channel models. It can also perform short-circuit evaluation of loss models for tasks like determining if receive power is above a certain threshold, thus speeding up simulations by up to an order of magnitude [12].

### VEINS INET Subproject

The Veins INET subproject[107] is licensed under GPL-2.0-or-later and adds using Veins as a mobility model through the INET Framework. The INET Framework is a GPL or LGPL licensed[108] open-source OMNeT++ model suite for wired, wireless, and mobile networks.

INET Framework features include 1) OSI layers (physical, link-layer, network, transport, application), 2) pluggable protocol implementations, 3) IPv4/IPv6 network stack, 4) transport layer protocols: TCP, UDP, SCTP, 5) routing protocols (ad-hoc and wired), 6) wired/wireless interfaces (Ethernet, PPP, IEEE 802.11, etc.), 7) physical layer with scalable level of detail (unit disc radio to detailed propagation models, frame level to bit/symbol level representation, etc.), 8) mobility support, 9) modeling of the physical environment (obstacles for radio propagation, etc.), and 10) visualization support[109].

Fig 11 shows an INET enhanced Veins simulation. Radio communication extensions are shown in green. A vehicle (e.g. car[0]) extended with a Veins_INET mobility module may have a customized app using UDP/IP to communicate with a wireless LAN system. The wireless LAN implemented in the INET Framework has a radio medium, physical environment, and visualizer.

---

[94] https://omnetpp.org/intro/
[95] https://eclipse.org/sumo
[96] https://github.com/eclipse/sumo/blob/master/LICENSE
[97] https://veins.car2x.org/features/
[98] https://www.x-plane.com/
[99] https://github.com/lsecicmc/AVENS/tree/Demonstration
[100] https://www.lsec.icmc.usp.br/en/avens
[101] https://veins.car2x.org/documentation/modules/#ieee80211p
[102] https://veins.car2x.org/documentation/modules/#obstacles
[103] https://veins.car2x.org/documentation/modules/#antennas
[104] https://veins.car2x.org/documentation/modules/#veins_gym
[105] https://veins.car2x.org/documentation/modules/
[106] https://veins.car2x.org/documentation/modules/#phy
[107] https://veins.car2x.org/documentation/modules/#veins_inet
[108] https://github.com/inet-framework/inet/blob/master/License
[109] https://inet.omnetpp.org/Introduction.html

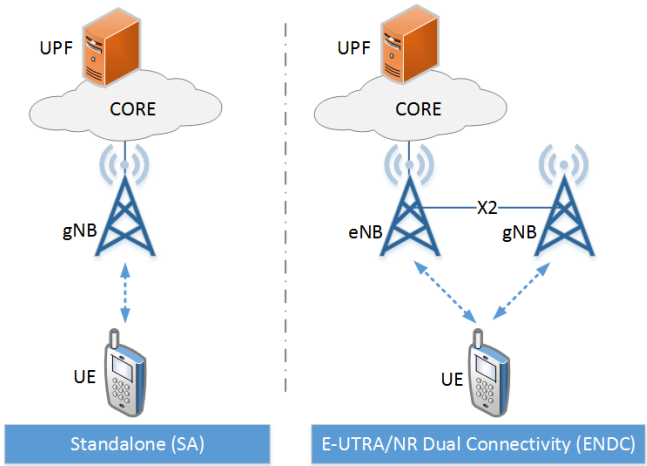

Fig. 12. Simu5G Standalone and E-UTRA/NR Dual Connectivity Simulation (ENDC)

The INET Framework has many showcase projects[110] available include: 1) Visualizing the Spectrum of Radio Signals[111], 2) Crosstalk between adjacent IEEE 802.11 channels[112], and 3) Coexistence of IEEE 802.11 and 802.15.4[113].

*Veins INET SimuLTE / 5G*

SimuLTE[114] is a simulation tool enabling complex system level performance-evaluation of LTE and LTE Advanced networks (3GPP Release 8 and beyond) for the OMNeT++ framework [13]. It is an LGPL v3[115] licensed C++ project available on github[116] building on top of OMNeT++ and INET Framework. SimuLTE also works with Veins, simulating LTE-capable vehicular networks.

SimuLTE models the data plane of the LTE/LTE-A Radio Access Network and Evolved Packet Core. It allows you to simulate LTE/LTE-A in Frequency Division Duplexing (FDD) mode, with heterogeneous eNBs (macro, micro, pico etc.), using both anisotropic and omnidirectional antennas. ENBs can be connected via X2 an interface and can communicate using both standard and user defined messages.

SimuLTE supports many features from applications down to the physical layer. Application support includes VoIP GSM AMR, H.264 video streaming, real-time gaming, and FTP. RLC support includes UM and AM segmentation and reassembly. MAC support includes buffering, PDU concatenation, CQI reception, transport format selection and resource allocation, and coding designed to facilitate cross-layer analysis. PHY features include transmit diversity using SINR curves, channel feedback computation, and realistic channel models. User terminal features include mobility, interference, all types of traffic, handover, and D2D communications. E-NodeB features include macro, micro, pico eNodeBs, inter-eNB coordination through X2 interface, handover, and CoMP, scheduling algorithms (Max C/I, Proportional Fair, Round Robin, etc.).

Simu5G is the evolution of the popular SimuLTE 4G network simulator incorporating 5G New Radio access [14][15]. Simu5G incorporates all ~~the~~ models from the INET library, which allows one to simulate generic TCP/IP networks including 5G NR layer-2 interfaces.

Fig 12 shows both standalone and dual connectivity Simu5G simulation. It simulates the data plane of the 5G RAN (rel. 16) and core network. It allows simulation of 5G communications in both Frequency Division Duplexing (FDD) and Time Division Duplexing (TDD) modes, with heterogeneous gNBs (macro, micro, pico, 5G Node Basestations), possibly communicating via the X2 interface to support handover and inter-cell interference coordination. Dual connectivity between an eNB (LTE base station) and a gNB is also available.

3GPP-compliant protocol layers are provided, whereas the physical layer is modeled via realistic, customizable channel models. Resource scheduling in both uplink and downlink directions is supported, including Carrier Aggregation and multiple numerologies, as specified by the 3GPP standard (3GPP TR 38.300, TR 38.211). Simu5G supports a large variety of models for mobility of UEs, including vehicular mobility.

Fig 11 shows Simu5G supports both StandAlone (SA) and E-UTRA/NR Dual Connectivity (ENDC) deployment where LTE and 5G coexist (3GPP – TR 38.801). In the ENDC configuration, the gNB works as a Secondary Node (SN) for an LTE eNB, which acts as Master Node (MN) and is connected to the Core Network. the LTE eNB model is imported from SimuLTE, with which Simu5G is fully compatible. The eNB and the gNB are connected through the X2 interface and all NR traffic needs to go through the eNB.

### 4.2 Other Wireless Communications Modeling

GNU Radio[117] is a GPLv3 licensed open-source software development toolkit with code available on github[118] providing signal processing blocks to implement software radios. It can be used with readily available low-cost external RF hardware to create software-defined radios, or without hardware in a simulation-like environment.

Open Air Interface is a BSD licensed implementation of the 3GPP specifications for the Evolved Packet Core Networks with code available on github[119]. Open Air Interface also provides Radio Access Network (RAN) code on gitlab[120] but under a special license not for commercial use[121].

srsRAN is a 4G/5G software radio suite developed by Software Radio Systems (SRS)[122]. It is available on github[123] with an AGPL license. Commercial license options are available from SRS[124]. Written in portable C/C++, the software includes: 1) srsUE, a full-stack SDR 4G/5G-NSA UE application with 5G-SA in development, 2) srsENB, a full-stack SDR 4G eNodeB application with 5G-NSA and 5G-SA in development, and 3) srsEPC, a light-weight 4G core network implementation with MME, HSS, and S/P-GW.

---

[110] https://inet.omnetpp.org/docs/showcases/
[111] https://inet.omnetpp.org/docs/showcases/visualizer/canvas/spectrum/doc/index.html
[112] https://inet.omnetpp.org/docs/showcases/wireless/crosstalk/doc/index.html
[113] https://inet.omnetpp.org/docs/showcases/wireless/coexistence/doc/index.html
[114] https://simulte.com/
[115] https://www.gnu.org/licenses/lgpl-3.0.en.html
[116] https://github.com/inet-framework/simulte
[117] https://www.gnuradio.org/about/
[118] https://github.com/gnuradio/gnuradio
[119] https://github.com/OPENAIRINTERFACE/openair-epc-fed
[120] https://gitlab.eurecom.fr/oai/openairinterface5g/
[121] https://www.openairinterface.org/?page_id=698
[122] https://www.srs.io/
[123] https://github.com/srsran/srsran

OpenLTE[125] is an AGPL licensed open-source implementation of the 3GPP LTE specifications with code available at sourceforge[126]. Octave code is available for test and simulation of downlink transmit and receive functionality and uplink PRACH transmit and receive functionality. In addition, GNU Radio applications are available for downlink transmit and receive to and from a file, downlink receive using rtl-sdr, HackRF, or USRP B2X0, LTE I/Q file recording using rtl-sdr, HackRF, or USRP B2X0, and a simple eNodeB using USRP B2X0. The current focus is on extending the capabilities of the GNU Radio applications and adding capabilities to the simple base station application (LTE_fdd_enodeb).

## 5 OPEN-SOURCE OPERATING SYSTEMS

### 5.1 Robot OS (ROS)

The Robot Operating System (ROS) is a flexible framework for writing robot software [7]. It is a collection of tools, libraries, and conventions aiming to simplify the task of creating complex and robust robot behavior across a wide variety of robotic platforms[127]. It is supported by the Open Robotics Foundation (as is Gazebo/Ignition) with code available under the permissive BSD3 license.

At the lowest level, ROS offers a message passing interface providing inter-process communication and is commonly referred to as a middleware. It uses a publish/subscribe system for messages that is anonymous and asynchronous. Data can be easily captured and replayed without any changes to code. Synchronous services are also available through preemptable Remote Procedure Calls (RPC).

ROS provides robot-specific features including a standard message format, a geometry library allowing many moving parts to be connected, and a universal robot description language that is not connected to Gazebo's simulation description format[128]. ROS provides built-in support for pose estimation, localization in a map, building a map, and mobile navigation. Debugging, plotting, and 3D visualization tools are also provided.

A very large number of robots use ROS spanning many categories of autonomous robots from manipulator arms, UGVs, commercial humanoids, and UASs[129]. Turtlebot, Roomba, and crazyflie (open-source open-hardware nano quadcopter) all use ROS.

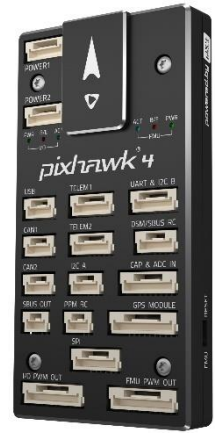

Fig. 13. Holybro Pixhawk 4

Both ArduPilot[130] and PX4 can be extended with ROS. MAVROS[131] is a BSD licensed MAVLink extendable communication node for ROS with proxy for Ground Control Stations[132]. Using MAVROS, ArduPilot can be extended with Simultaneous Localization And Mapping (SLAM) for non-GPS navigation, object avoidance, and SITL with Gazebo[133]. Similarly, PX4[134] supports ROS 2 over a PX4-ROS 2 bridge[135] using Real-Time Publish Subscribe (RTPS)[136] or to ROS 1 using MAVROS[137].

ROS 2 is the newest version of ROS[138]. Many tutorials are available with step-by-step instructions[139]. A step-by-step demo of using turtlebot 3 with ROS and Gazebo is provided by Ubuntu[140].

## 6 OPEN-SOURCE HARDWARE

Open-source hardware is hardware whose design is publicly available so that anyone can study, modify, dis-tribute, make, and sell the design or hardware based on that design.

### 6.1 Pixhawk

Pixhawk provides open standards for drone hardware[141]. It is designed for manufacturers wanting to build PX4 compatible products[142]. Standards include the Pixhawk Autopilot Reference Standard, the Pixhawk Payload Bus Standard, and the Pixhawk Smart Battery Standard.

Many qualified Pixhawk standard designs exist. Fig. 13 shows the Holybro Pixhawk 4 - an advanced autopilot designed and made in collaboration with Holybro and the PX4 team[143]. Conforming to the FMUv5 specification, it includes: a 32-bit ARM Cortex-M7 processor running at 216Mhz to execute the FMUv5 open hardware design, a 32-bit ARM Cortex-M3 running at 24Mhz for I/O, several sensors including accelerometer, gyroscope, magnetometer,

---

barometer, and u-blox GPS. A key feature of Pixhawk 4 is firmware and QGroundControl integration. The PX4 project uses Pixhawk standard autopilots and directly supports the Holybro Pixwak 4 hardware[144].

Pixhawk autopilot standards include pin-out definition and references, architectural block diagram, PCB layout guidelines, and connector specifications[145].

Fig 14. Shows the Pixhawk FMUv6X system design. While still under draft at the time of writing, it has the same architecture as the v5X but is based on the STM32H7 chip[146].

# 7 CONCLUSIONS

Unmanned aerial systems can be developed using open-source software and hardware. Each layer of a top-down hierarchy typically has multiple choices for open-source deployment. Every aspect of traffic management, flying of the aircraft, simulation of the environment, simulation of UAV interactions, to electronic hardware controlling flight has open-source solutions. In this brief survey we describe some popular open-source hardware and software for realizing unmanned aerial systems.

## ACKNOWLEDGMENT

The authors wish to thank the members of the Wireless Innovation Forum's Unmanned Vehicle Wireless Networking Special Interest Group for their valuable feedback on open-source projects.

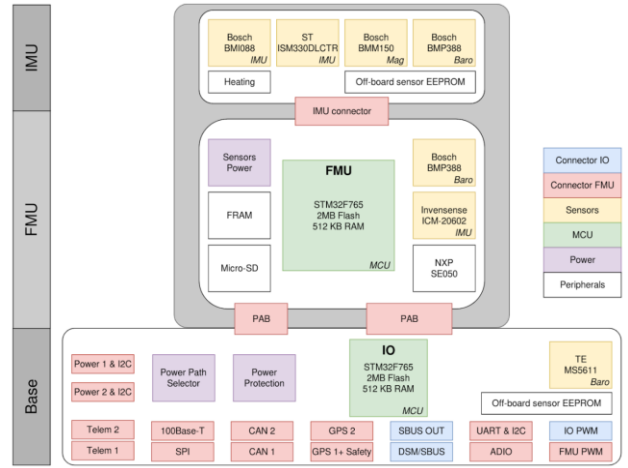

Fig. 14. Pixhawk FMUv6X Overview

**John Glossner** is CEO of Optimum Semiconductor Technologies Inc. (OST) d/b/a General Processor Technologies (GPT), and CTO of Hua Xia General Processor Technologies, the China-based parent company of OST. OST designs heterogeneous systems based on its Unity architecture supporting CPU, GPU, DSP, and AI processors. Previously he co-founded Sandbridge Technologies and served as EVP & CTO. Prior to Sandbridge, he managed both technical and business activities in DSP and Broadband Communications at IBM and Lucent/Motorola's Starcore. He also serves as Chair of both the Wireless Innovation Forum and the Heterogeneous System Architecture Foundation. He is also the chair of the Computer Architecture, Heterogeneous Computing, and AI Lab at the University of Science and Technology Beijing. He received a Ph.D. in Electrical Engineering from TU Delft in the Netherlands, M.S. degrees in E.E. and Eng.Mgt. from NTU and holds a B.S.E.E. degree from Penn State. John has more than 120 publications and 45 issued patents.

**Samantha Murphy** Samantha Murphy earned her M.B.A. and B.A. degrees from Southern New Hampshire University and currently serves as a project manager for Optimum Semiconductor Technologies Inc. (OST) d/b/a General Processor Technologies (GPT), a subsidiary of China-based Hua Xia General Processor Technologies. OST designs heterogeneous systems based on its Unity architecture supporting CPU, GPU, DSP, and AI processors. Samantha oversees several projects in OST's program of smart home technologies and has multiple international patents pending.

---

[144] https://docs.px4.io/master/en/flight_controller/pixhawk_series.html
[145] https://pixhawk.org/standards/
[146] https://www.st.com/en/microcontrollers-microprocessors/stm32h7-series.html

**Daniel Iancu** earned his MS and Ph.D. degree in physics from the University Babes-Bolyai, Cluj-Napoca, Romania. Currently, he is the director of technology at OST Inc. Prior to OST he held a teaching position at the University Babes-Bolyai, Cluj-Napoca, Romania and several positions in US: R&D engineer at Medical Laboratory Automation, Director of R&D at Tempo Inc, Principal Engineer at Pitney Bowes Research, Systems Manager at Com21, Director of communication and Technology at Sandbridge. He has over 60 published papers and over 40 US and international patents.